\documentclass[letterpaper, 10 pt, conference]{ieeeconf}  

\IEEEoverridecommandlockouts                              

\usepackage{amsmath,amssymb,amsfonts}
\usepackage{algorithmic}
\usepackage{graphicx}
\usepackage{textcomp}
\usepackage[table]{xcolor}
\usepackage{caption}
\usepackage{multirow}
\usepackage{footnote}
\usepackage{hyperref}
\usepackage{array}
\usepackage{amssymb}
\usepackage{todonotes}
\usepackage[pscoord]{eso-pic}
\usepackage{multirow}
\usepackage{mathtools}
\usepackage{cite}
\usepackage{algorithm2e}

\title{\LARGE \bf
Leaving the Lines Behind: Vision-Based Crop Row Exit for Agricultural Robot Navigation
}

\author{Rajitha de Silva$^{1}$, Grzegorz Cielniak$^{2}$ and Junfeng Gao$^{3}$
\thanks{This work was supported by Lincoln Agri-Robotics as part of the Expanding Excellence in England (E3) Programme. }
\thanks{$^{1}$Rajitha de Silva, $^{2}$Grzegorz Cielniak and $^{3}$Junfeng Gao are with Lincoln Agri-Robotics Centre, Lincoln Institute for Agri-Food Technology, University of Lincoln, UK
        {\tt\small $^{1}$rajitha@ieee.org, $^{2}$gcielniak@lincoln.ac.uk, $^{3}$jugao@lincoln.ac.uk} (Correspondence author: Junfeng Gao)}%
}

\begin{document}

\maketitle
\begin{abstract}
Usage of purely vision based solutions for row switching is not well explored in existing vision based crop row navigation frameworks. This method only uses RGB images for local feature matching based visual feedback to exit crop row. Depth images were used at crop row end to estimate the navigation distance within headland. The algorithm was tested on diverse headland areas with soil and vegetation. The proposed method could reach the end of the crop row and then navigate into the headland completely leaving behind the crop row with an error margin of 50 cm. 
\end{abstract}

\section{Introduction}

Agricultural robots are emerging as an indispensable tool in modern farming as a solution for rising labour demands in the sector. This growing demand for agricultural robots has prompted developers to seek efficient ways to reduce production costs for these technologies. Computer vision technologies provide an appealing avenue for cost reduction in agri-robots \cite{wang2022applications}. Vision based navigation in agri-robots provide the option to get rid of expensive and unreliable Real Time Kinematic Global Positioning System (RTK-GPS) and Lidar systems\cite{ball2016vision, gil2023low}. 

Crop row following algorithms are often used to navigate arable fields for precision agriculture. Most of such systems emphasise on detecting crop rows and following them based on vision input \cite{bonadies2019overview}. Switching from one crop row to another often remains unsolved using computer vision\cite{winterhalter2021localization}. Most of the existing vision-based navigation schemes would rely on some form of GPS aid to switch from one row to another\cite{ahmadi2020visual}. The systems that present a vision-based solution require specific hardware requirements such as bidirectional camera setup with the symmetric operation of the robot\cite{ahmadi2022bonnbot}. A vision system that provides both crop row following and row switching using a single front-mounted camera would enable vision-only navigation schemes for agri-robot navigation in arable fields. 

The problem of row switching could be decomposed as three tasks namely, T1: Exiting the current crop row, T2: Turning around towards the field, and T3: Re-entering the next crop row. The goal of this paper is to provide a vision-based solution for T1 task. The T1 task is critical to all the robot configurations while T2 and T3 tasks are highly dependent on the specific configuration of the robot. The T2 task could be solved using most existing solutions such as sliding mode control\cite{tu2019robust}. Perception part of task T3 could be framed as an image correspondence problem in computer vision. The control problem of T3 could be realised using path planning algorithms \cite{patle2019review, peng2022depth}. The main contribution of this paper is a purely vision based crop row exit pipeline. The proposed pipeline was evaluated against odometry based navigation baseline.

\section{Methodology}
\label{sec:mtd}
The task T1 should be initiated when the robot is nearing the end of the currently traversing crop row. The End Of Row (EOR) detector from our previous work \cite{de2022vision} is used to trigger the T1 task. The EOR detector tracks the row end of a crop field and triggers a signal when the row end reaches the middle of the camera frame. 

\begin{figure}
\centering
\includegraphics[scale=0.25]{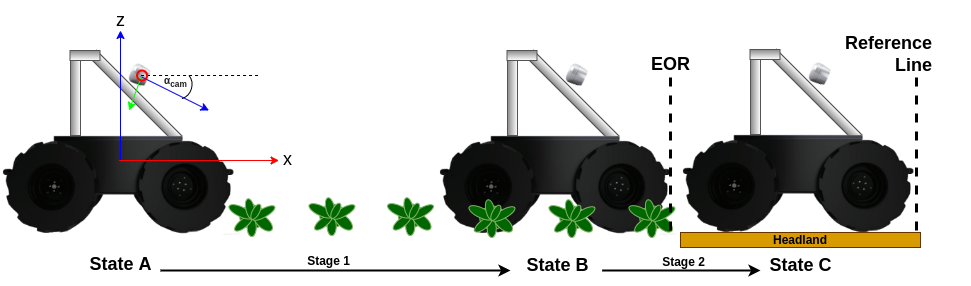}
\caption{Traversal States and Stages of Completion for T1 Task}
\label{fig:stt}
\end{figure}

T1 task is divided into two stages. Stage 1 attempts to drive the robot from EOR detection state (State A) to EOR position (State B). The robot is driven from EOR position to the headland position (State C) where the robot traverses a distance $L$ where $L=m\times l$. The length of the robot is $l$ and $m$ is a scale factor ($m=1$ in our experiment). The states A, B and C are illustrated in Figure \ref{fig:stt}. Both stage 1 and stage 2 use local image feature matcher for the respective navigation. 

\subsection{Local Feature Similarity Matcher (LFSM)}
\label{ssec:sift}
The LFSM calculates the local image feature descriptors in two images. Scale-Invariant Feature Transform (SIFT) \cite{lowe2004distinctive} was used to calculate feature descriptors in this implementation. The first image is captured at a specific traversal state (State A or State B) and the second image is captured while the robot is traversing from one state to the next. The k- Nearest Neighbour\cite{cunningham2021k} algorithm (KNN) was used to find matches between the resulting descriptors with a k value of 2. The matches from knn algorithm are further filtered if the ratio of distances between the first and second match is below $0.7$. 

\subsection{Navigation Stage 1}
\label{ssec:s1}
The navigation stage 1 is initiated by the EOR detection trigger. The bottom part of the image captured at state A is cropped below the EOR position in image space. This cropped image is used as $image1$ for LFSM. The consecutive images (for $image2$ in LFSM) were captured while traversing forward in the crop row with a fixed forward linear velocity. Each of these images were also cropped using the same mask used to crop $image1$ based on EOR potition in image space. The robot is halted when the $sim\_score$ of LFSM algorithm falls below a threshold (experimentally determined: $20$).

\subsection{Navigation Stage 2}
\label{ssec:s2}
\begin{figure}
\centering
\includegraphics[scale=0.27]{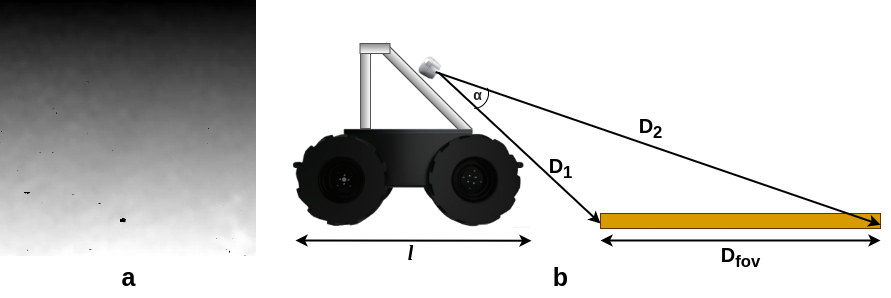}
\caption{Navigation Stage 2 Cropping Mask Parameters}
\label{fig:ns2}
\end{figure}

Navigation stage 2 is started after the stage 1 where another image is captured to be used as $image1$ for LFSM. It was assumed that this image would only contain the ground space of headland area without any tall obstacles. The depth frame corresponding to this image would look like an increasing gradient (bottom to top) despite some noise from ground imperfections and weeds as seen in Figure \ref{fig:ns2}.a. The distance along the headland seen in camera frame is defined as $D_{fov}$. The median value of the pixels corresponding to the top and bottom rows of this depth image will be used as distances $D_1$ and $D_2$ to calculate $D_{fov}$ as indicated in Figure \ref{fig:ns2}.b. 

The  $D_{fov}$ is calculated using law of cosines ($D_{fov} = \sqrt{D_1^2 + D_2^2 - 2 D_1 D_2 \cos(\alpha)}$) where $\alpha$ is the field of view angle for the camera (58\textdegree for Realsense D435i). The ratio between $D_{fov}$ and length of the robot ($l$) was used to determine the cropping mask for $image1$ in stage 2. This cropping mask will represent the part of the RGB image corresponding to a distance equivalent to $l$. The LFSM algorithm is executed while the robot is moving forward with a fixed linear velocity until the similarity score falls below the threshold (experimentally determined: $20$). The robot is then halted completing the T1 task. 

The navigation stages 1 and 2 uses LFSM based visual feedback to exit crop row. Depth data is used at the beginning of stage 2 to estimate the traversal distance.

\section{Results and Discussion}
\label{sec:rslt}
\begin{figure}
\centering
\includegraphics[scale=0.4]{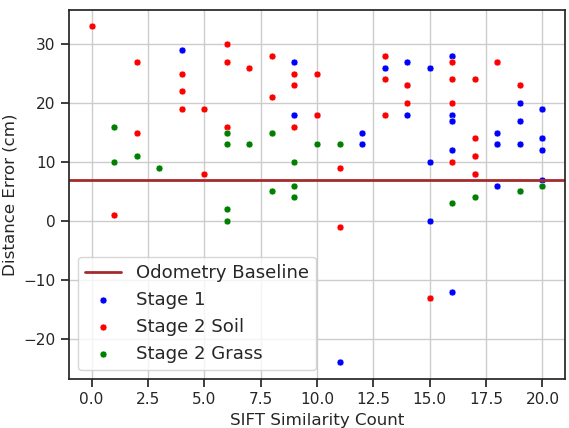}
\caption{Row Exit Distance Error for Task T1 (Positive error when robot traverse past the desired position)}
\label{fig:t1}
\end{figure}

An experimental crop field was setup using plastic plants placed on a real field for controlled yet realistic scenario. The robot was positioned at random distances ($<$1m) away from state A and allowed to follow the crop row until EOR detection triggers, launching navigation stage 1. The distance from the frontmost position of the robot to the real EOR position was measured during the experiment. The results from 40 stage 1 trails are displayed with blue markers in Figure \ref{fig:t1}. The median absolute error for stage 1 trials was found to be 24 cm which is less than the distance of 2 plants in the crop row. The experiment yields a net positive error since 92.5\% of the trials had a positive error.

A baseline experiment was conducted to determine the ability of the robot to complete stage 2 only using odometry data. The median error for the baseline experiment was 7cm. During the testing of stage 2, the system was evaluated in two distinct headland areas, namely, an area with soil and another with vegetation. The median absolute error in soil headland was 21 cm and the verdant headland was 9.5 cm. Verdant headlands perform better due to distinct image features. 

Although the forward linear velocity of the robot was fixed in both experiments, it was observed that the performance could be improved by reducing the speed of the robot. The proposed pipeline runs at a 1.5 frames per second rate. A slower velocity would result in shorter gaps between consecutive images and hence early detection of state transition. 

\section{Conclusion}
\label{sec:con}
The study has demonstrated that the proposed method is a promising approach for exiting crop rows in arable fields for purely vision based crop row navigation. Vision based navigation is most suitable for stage 1 navigation where target distance is dependent on the camera pitch angle. Vision only navigation in stage 2 performs inferior to odometry baseline in the absence of robust features. Odometry based navigation is more suitable for stage 2 since the target distance for navigation is fixed. The robot could exit the crop row with an overall average error of 24 cm during the combined(stages 1 and 2) vision only navigation experiments. It was also noted that the verdant headlands result in accurate navigation in stage 2 due to the abundance of features. Headlands with lengths shorter than the length of the robot cannot be navigated using the proposed method.

\bibliography{root}
\end{document}